\documentclass[letterpaper, 10 pt, conference]{ieeeconf}  
\IEEEoverridecommandlockouts                              
\usepackage{graphicx}
    \graphicspath{{images/}}
\usepackage{mathptmx}
\usepackage{stmaryrd}
\usepackage{amsmath}
\usepackage{amssymb}
\usepackage{mathrsfs}
\usepackage{xcolor}
\usepackage{hyperref}
\usepackage{svg}
\usepackage[ruled,vlined]{algorithm2e}
\usepackage[export]{adjustbox}
\usepackage{comment}
\newenvironment{claim}[1]{\par\noindent\underline{Claim:}\space#1}{}
\newenvironment{claimproof}[1]{\par\noindent\underline{Proof:}\space#1}{\hfill $\square$}

\usepackage[]{cite}

\title{\LARGE \bf
Signal-based self-organization of a chain of UAVs\\ for subterranean exploration
}

\author{
    Pierre Laclau$^{1,2}$, Vladislav Tempez$^{2}$, Franck Ruffier$^3$, Enrico Natalizio$^{4}$, Jean-Baptiste Mouret$^{2}$%
    \thanks{*This work received funding from the European Research Council (ERC) under the European Union’s Horizon 2020 research and innovation programme (GA no. 637972, project “ResiBots”) and the ANR/DGA ASTRID project ``Proxilearn''.}
    \thanks{$^{1}$ Université de Technologie de Compiègne (UTC), France}
    \thanks{$^{2}$ Inria, CNRS, Universit\'e de Lorraine, LORIA, Nancy F-54000, France}
    \thanks{$^3$ Aix Marseille Universit\'e, CNRS, ISM, Marseille F-13009, France}
    \thanks{$^{4}$ Universit\'e de Lorraine, CNRS, LORIA, Nancy F-54000, France}
    \thanks{Corresponding author: \href{mailto:jean-baptiste.mouret@inria.fr}{jean-baptiste.mouret@inria.fr}}
   }

\begin{document}

\maketitle
\thispagestyle{empty}
\pagestyle{empty}

\begin{abstract}
Miniature multi-rotors are promising robots for navigating subterranean networks, but maintaining a radio connection underground is challenging. In this paper, we introduce a distributed algorithm, called U-Chain (for Underground-chain), that coordinates a chain of flying robots between an exploration drone and an operator. Our algorithm only uses the measurement of the signal quality between two successive robots as well as an estimate of the ground speed based on an optic flow sensor. We evaluate our approach formally and in simulation, and we describe experimental results with a chain of 3 real miniature quadrotors (12 by 12 cm) and a base station.
\end{abstract}

\section{INTRODUCTION}

Thousands of subterranean networks permeate the underground: caves, utility tunnels, abandoned  mines, underground quarries, sewers, etc. These voids often need to be mapped and inspected, typically to ensure the safety of new buildings or tunnels, but also in case of obstruction or intrusions.


Robots would greatly help the inspection of these networks, which are often too confined for humans (e.g., sewer pipes) or too dangerous (caves, abandoned mines, collapsed buildings) \cite{morris2006recent}. However, designing such robots is challenging. Firstly, they need very good off-road abilities, as the floor is typically uneven, with steep inclines, and sometimes flooded. Secondly, they have to be small enough to fit into tunnels that are often too narrow for humans. Overall, ground robots (wheels or tracks) are either too small to clear large obstacles or too big to enter narrow voids.

Flying robots (Unmanned Aerial Vehicles -- UAVs) are a promising alternative to ground robots for subterranean operations \cite{rogers2017distributed,freire2017capture,preston2017use,mansouri2019autonomous,1952_24_Jones}. As they fly, they are not impaired by obstacles or liquid on the floor. They can also easily fly over steep inclines, stairs, and even ladders. In addition, current quadrotors are cheap, well understood, and miniaturized. For instance, the Crazyflie is a research quadrotor that fits in a 12 $\times$ 12 cm square (rotors included) and weights less than 40g~\cite{giernacki2017crazyflie}.

\begin{figure}
    \centering
    \includegraphics[width=1\linewidth]{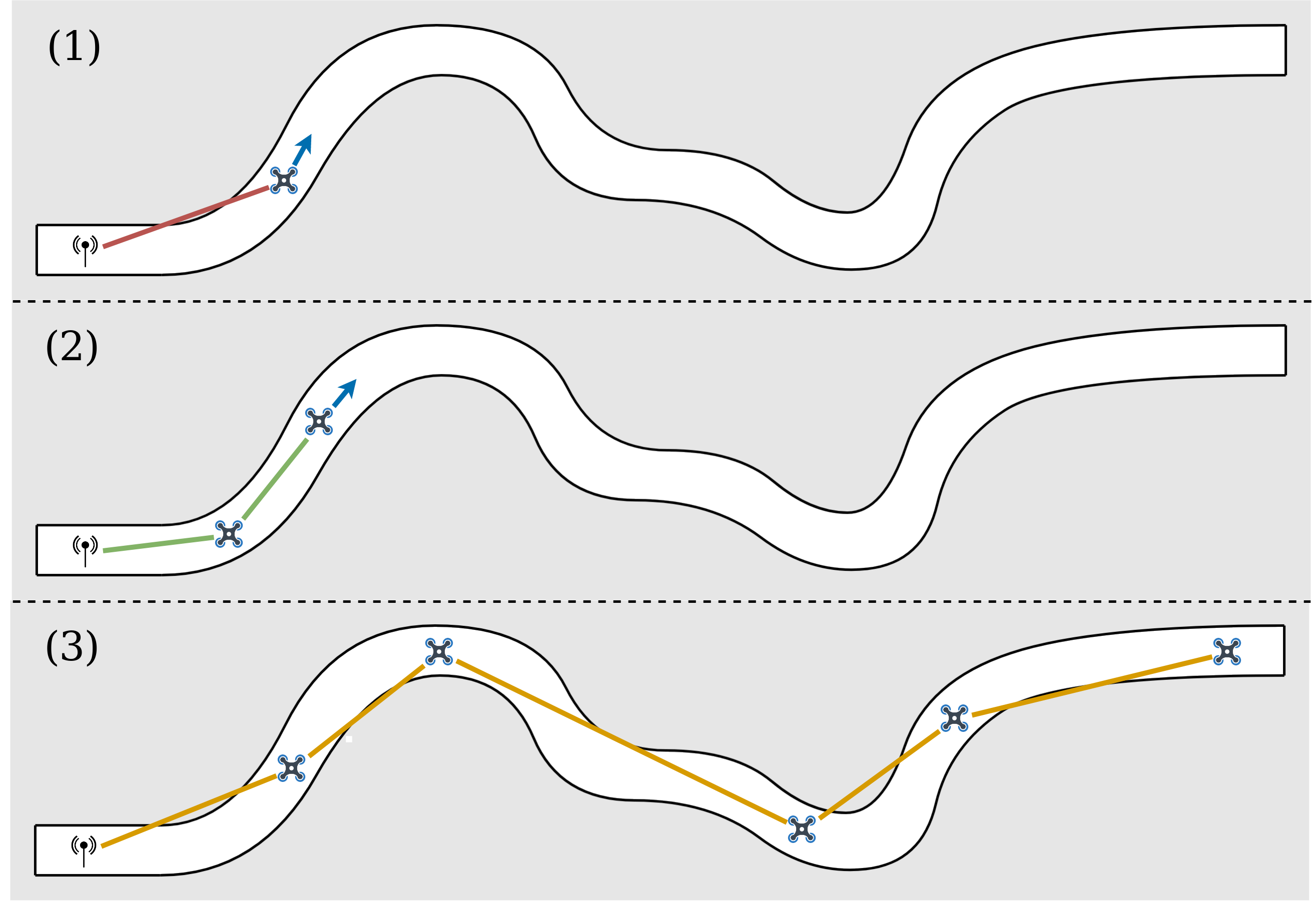}
    \caption{\textbf{Illustration of the U-chain algorithm to maintain a communication chain in a tunnel}. A human operator controls an explorer UAV in a tunnel. When the signal strength becomes too weak $(1)$, a new autonomous UAV takes off and acts as a relay by finding the best relay position
        $(2)$. The objective is the exploration of an environment at any distance, while maximising the global signal quality $(3)$.}
    \label{fig:example_tunnel}
    \vspace{-1em}
\end{figure}

Unfortunately, small flying robots cannot carry a cable to stay connected with the operator, contrary to ground robots. Instead, they have to rely on a radio link, which works well in open air environments but represents a big challenge in subterranean environments \cite{khuwaja_survey_2018,akyildiz_signal_2009}.
In particular, a typical signal such as the Crazyflie's 2.4GHz radio miniature system
does not penetrate large amount of rocks and even bounces in metallic pipes. This means that the radio works underground only when there exists an unobstructed path between the emitter and the receiver: at each turn of a corridor, the signal is severely deteriorated or lost.

In this paper, we introduce a distributed algorithm, called U-Chain (for Underground-chain), that coordinates a chain of UAVs, so that UAVs act as relays between an exploration drone and an operator. Our algorithm assumes that the UAVs are flying in a tunnel, but it only uses the measurement of the signal quality between each couple of UAVs and an estimate of the ground velocity based on an optic flow sensor. It does not need any localization system (e.g., a visual SLAM algorithm) and, as such, (1) it naturally adapts to the material and the topography of the tunnel and (2) it works on miniature platforms with limited computational power. For instance, our algorithm will place a relay at a corner (without knowing that it is a corner), but it will also increase the number of relays if the signal is highly perturbed in a specific zone.

We show formally that maximizing the signal quality of a communication chain in a tunnel is equivalent to equalizing the signal quality between the UAVs, which leads to a simple movement policy for the UAVs. Nonetheless, the signal quality depends on many parameters besides the distance between two UAVS, which results in a noisy value. To address this challenge, we introduce a Kalman filter that leverages optic flow-based measurements of the ground speed. We study the algorithm extensively in simulation and provide experimental results on a chain of 3 miniature quadrotors and a base station.





\section{BACKGROUND}
Most of the work about subterranean exploration has been focused on a single wheeled or tracked robot \cite{morris2006recent} that is coupled with a SLAM algorithm to build a 3-dimensional map \cite{thrun2004autonomous}. Single UAVs are, however, increasingly tested in mines and they show promising results \cite{freire2017capture,preston2017use,mansouri2019autonomous,1952_24_Jones}. They have, for instance, been deployed by many teams during the DARPA Subterranean Challenge (\url{https://www.subtchallenge.com/}, 2020).

In parallel, many algorithms have been proposed for exploring unknown environments with a group of robots, with or without maintaining a permanent connection, and with or without a global coordination~\cite{amigoni_multirobot_2017,dorigo2013swarmanoid}. To our knowledge, the vast majority of algorithms assume that the robots explicitly estimate their absolute or relative positions \cite{amigoni_multirobot_2017,pei2010coordinated,cesare2015multi,stump2011visibility,nestmeyer2015decentralized}, which they can share with the other robots, and sometimes know the environment beforehand \cite{stump2011visibility, hsieh2008maintaining}. Using a positioning system makes sense in outdoor situations, in which the GPS signal is available, and indoor when each robot embeds a SLAM algorithm. Nevertheless, underground miniature UAVs cannot use GPS and they do not have the computational power and sensors needed for accurate SLAM. 
A notable exception for outdoor networks is the work of Hauert \emph{et al.} \cite{hauert_evolved_2009}, which leverages an evolutionary algorithm to find a reactive strategy that maintains the network without positioning information.

Signal quality estimates are often considered too unreliable to solely drive the organization of the group of robots \cite{amigoni_multirobot_2017}, in particular because it is influenced by many factors that are unknown from the robots and therefore not modeled \cite{amigoni_multirobot_2017,khuwaja_survey_2018}. Nonetheless, they were recently used in an exploration algorithm for the Crazyflie quadrotors (the same quadrotors as those used in the present work) \cite{mcguire2019minimal} for three purposes: (1) going back to the base station (by following the gradient of signal quality between the quadrotor and the base), (2) avoiding the other robots (by looking at the inter-quadrotor signal quality), and (3) adapting the exploration direction (by choosing directions that move away from the other quadrotors). This latter system shows that the signal quality can be useful, but it assumes that the base station can always communicate with the UAVs: the objective is not to maintain the communication, but to explore with a lightweight strategy based on reactive behaviors and a state machine.

In the present work, robots are assumed to be in a tunnel, which means they only need to decide if they go forward or backward to maintain the connectivity: this simplifies the coordination problem. In that particular situation, we hypothetize that the signal quality estimates can be sufficient to maintain a chain of communication. This kind of signal-based coordination was recently investigated in the particular case of subterranean tunnels by Rizzo \emph{et al.} with wheeled robots \cite{rizzo_signal-based_2013}. In their article, the authors analyze experimentally and theoretically the signal propagation in tunnels to obtain the general characteristic parameters; they describe a technique that uses these parameters to coordinate robots. While this approach is successful, it assumes a precise identification of the parameters of the radio signal before deployment, which we cannot obtain while exploring complex environments. Moreover, the approach of Rizzo \emph{et al.} slightly differs from ours in their objective: while they aim at keeping the quality greater than a minimum threshold, our algorithm maximizes the global quality of the connection.

\section{PROBLEM FORMULATION} \label{sec:problem_formulation}

We consider a chain of UAVs in a corridor without any positioning system. The general objective is to maintain a high-quality connection between each UAV and the next so that the first UAV of the chain can communicate with the last one. As the UAVs are in a corridor, they only need to decide whether to move forward or backward; they base their decision on the estimates of the signal quality (RSSI) with the previous UAV and with the next UAV in the chain. 

Since the network is organized as a linear formation, a good way of estimating the global signal strength is by finding the relay link with the lowest chances of successfully transmitting the packet. We will call it the worst \emph{bottleneck} in the communication chain.
In this problem, this corresponds to maximizing the worst signal quality between two consecutive UAVs in the communication chain.

\subsection{Notations}
\begin{itemize}
    \item $\mathcal{C}$ 2D curve that describes the corridor geometry.
    \item $A=\{a_0,\dots,a_n\}$ set of UAVs in the 
    chain.
    \item $R \subset A,\, R=\{a_1,\dots,a_{n-1}\}$ autonomous relay drones that physically position themselves between the head $H=a_0$ and the base $B = a_n$ to create a relay chain.
    \item $s_{min}$ manually set threshold indicating whether the signal strength ensures a stable connection or not. 
    \item $P=\{x_0,\dots,x_n\}$ set of curvilinear abscissa positions of the UAVs along $\mathcal{C}$ starting at 0.
    \item $s(x_1,x_2)$ the function giving signal quality between the positions $x_1$ and $x_2$.
\end{itemize}

\subsection{Assumptions}
\begin{itemize}
    \item All agents position themselves at the center of the tunnel.
    \item $\forall i,j \in \llbracket 0, n \rrbracket^2,\, i>j, \, x_i<x_j$: all drones are positioned in the chain according to their index.
    \item The function $(x_1,x_2) \mapsto s(x_1,x_2)$ is defined only for $x_1<x_2$ and is continuous, i.e. we expect by convention to have the position closer to the origin as first argument of $s$. A symmetric extension of $s$ could be written as $\hat{s}: (x_1,x_2) \mapsto s(min(x_1,x_2), max(x_1,x_2))$.
    \item $(x_1,x_2) \mapsto s(x_1,x_2)$ is a decreasing function: if $[x_1, x_2] \subset [\Tilde{x}_1, \Tilde{x}_2]$ then $s(x_1,x_2)>s(\Tilde{x}_1,\Tilde{x}_2)$ (that is, the signal quality decreases when UAVs move away from each other).
\end{itemize}


\subsection{Objective function} \label{sec:objfun}
The optimal configuration is the solution of:
\begin{equation}
    x^* = \operatornamewithlimits{argmax}_{x_i} \left(\min_{i}\left( s(x_{i+1}, x_i) \right)\right)
    \label{eq:first_cost}
\end{equation}
that is, we search for the set of positions that maximizes the signal quality of the weakest link.
\begin{claim}
    If the function $s$ is continuous then the optimal configuration verifies:
    \begin{equation}
        \forall ~ (i,j) \in \llbracket 0,n-1 \rrbracket \quad s(x_{i+1}, x_{i}) = s(x_{j+1}, x_{j})
        \label{eq:cost_equal}
    \end{equation}
\end{claim}
\begin{claimproof}
    \label{proof:equality_constraints}
    (by contradiction). We assume that there exists an optimal solution $x^*$ that does not verify (\ref{eq:cost_equal}) (i.e., $\exists i,~ s(x^*_{i+1}, x^*_{i}) < s(x^*_{i},x^*_{i-1})$, or $s(x^*_{i+1}, x^*_{i}) < s(x^*_{i+2},x^*_{i+1})$). $x^*$ can be improved by moving the UAV $a_i$ (the one at the junction of the two unequal links at position $x_i$) by a step $\epsilon$ in the direction that improves the weakest of the two links. 
    Because of the continuity of $s$, there exists necessarily an $\epsilon$ that is small enough so that the previously strongest link is deteriorated to a value that is still better than the previously weakest link. This new configuration is better than the optimal one, giving the contradiction.
\end{claimproof}

These equality constraints (equation~\ref{eq:cost_equal}) define a necessary condition for the optimal configuration that we will later prove to also be sufficient. Please note that the fact that all the links' qualities are equal does not imply that the distance between two successive UAVs is equal, because the link quality is affected by the environment (e.g., a turn or a different wall material).

We denote by $s_{eq}$ the link quality of configurations with equal links. We know that at least one exists because the algorithm introduced in Section \ref{sec:algorithm} is proven to converge to equal links' qualities.

\begin{claim}
    Given the initial and end positions of the chain ($x_0 = x_t$ and $x_n=0$) and the number of UAVs, if the function $s$ is a decreasing function (i.e. $[x_1, x_2] \subset [\Tilde{x}_1, \Tilde{x}_2] ~ \Rightarrow ~ s(x_1,x_2)>s(\Tilde{x}_1,\Tilde{x}_2)$) then any configuration with equal links is a solution of the optimization problem stated in (\ref{eq:first_cost}).
\end{claim}

This result is not straightforward as we do not know much about the signal quality function $s$. 
In particular, it is straightforward to find examples of non optimal solutions with equal links' qualities when $s$ is \emph{not} decreasing. For example, when the UAVs are in a u-shaped environment and the signal can traverse the walls, the optimal signal quality would be achieved by putting all the UAVs at the end of the U (closest to the starting point when we ignore the walls), but there are many sub-optimal solutions with equal signal's qualities but a worst global quality.

\begin{claimproof}
    (by contradiction). We assume that there exist at least two configurations $C$ and $\Tilde{C}$ with all equal links but a different signal quality for these links and same fixed position for first and last UAVs i.e.:
    \noindent\begin{minipage}{.5\linewidth}
        \begin{equation}
            \begin{cases}
                \forall i<n ~ s(x_{i+1}, x_i) = s_{eq} \\
                x_0 = x_t                                                      \\
                x_n = 0
                \label{eq:constraints_C}
            \end{cases}
        \end{equation}
    \end{minipage}%
    \begin{minipage}{.5\linewidth}
        \begin{equation}
            \begin{cases}
                \forall i<n ~ s(\Tilde{x}_{i+1}, \Tilde{x_i}) = \Tilde{s_{eq}} \\
                \Tilde{x}_0 = x_t                                                              \\
                \Tilde{x}_n = 0
                \label{eq:constraints_C_Tilda}
            \end{cases}
        \end{equation}
    \end{minipage}
    \begin{equation}
        \textrm{and: } \Tilde{s_{eq}} < s_{eq}
        \label{eq:inequal_signal_quality}
    \end{equation}

    We prove by recurrence that all $\Tilde{C}$ UAVs positions $\Tilde{x}_i$ are farther than those of $C$: $\forall i < n$, $\Tilde{x}_i > x_i$.

    We have $x_n = \Tilde{x}_n$ because the last UAV is fixed. As a consequence we have either
    \begin{equation}
        [x_n, x_{n-1}] \subset [\Tilde{x}_n, \Tilde{x}_{n-1}]\\
        ~\textrm{or}~
        [\Tilde{x}_n, \Tilde{x}_{n-1}] \subseteq [x_n, x_{n-1}].
    \end{equation}
    If the last option was true, because $s$ is decreasing, we would have:
    \begin{equation}
        \Tilde{s}_{eq} = s(\Tilde{x_n}, \Tilde{x}_{n-1}) \geq s(x_n, x_{n-1}) = s_{eq}
    \end{equation}
    that in contradiction with (\ref{eq:inequal_signal_quality}), therefore $x_{n-1} < \Tilde{x}_{n-1}$.

    Now, if we assume that $\Tilde{x}_{i+1} >x_{i+1}$ we cannot have $\Tilde{x}_{i} \leq x_{i}$ as it would give that
    $[\Tilde{x}_{i+1}, \Tilde{x}_i] \subset [x_{i+1}, x_i]$
    and thus 
    $\Tilde{s}_{eq} = s(\Tilde{x}_{i+1}, \Tilde{x}_i) > s(x_{i+1}, x_i) = s_{eq}$,
    which is in contradiction with (\ref{eq:inequal_signal_quality}), therefore
    $
        (\Tilde{x}_{i+1} >x_{i+1}) \Rightarrow (\Tilde{x}_{i} >x_{i})
    $.
    This proves by recurrence that $\Tilde{x}_0 > x_0$ i.e. that $C$ and $\Tilde{C}$ have different position for the first UAV, which is in contradiction with the initial hypotheses.
\end{claimproof}

Moreover in this proof we showed that signal quality determines UAVs positions given position of first and last UAVs.

Till now, the optimization problem has been defined for a fixed number of UAVs. We also want to use the smallest possible number of UAVs to explore a given area. To do that, we modify the optimization problem to minimize the number of UAVs, keeping the signal quality equal between drones:

\begin{align}
    \label{eq:cost_n_drones}
&n^* =  \operatornamewithlimits{argmin} n\\
&\textrm{subject to: }\\
&    \begin{cases}
        \forall i,j \in \llbracket 0, n-1 \rrbracket^2 \quad s(x_{i+1}, x_i)=s(x_{j+1}, x_j)       \\
        \forall i \in \llbracket 0, n-1 \rrbracket \quad s(x_{i+1}, x_i) \geq s_{\min} \\
        x_n = 0                                                            \\
        x_0 = x_t
    \end{cases}
    \label{eq:constraints_final}
\end{align}
Enforcing the equality of all links' qualities ensures that we still optimize the worst signal quality but here we also constrain to a minimal signal quality $s_{min}$ to prevent signal loss along the chain.

\section{POSITIONING ALGORITHM} \label{sec:algorithm} 
Our objective is to design a distributed algorithm that minimizes the number of UAV (equation~\ref{eq:cost_n_drones}) under the constraints defined by (\ref{eq:constraints_final}).

\subsection{Convergence to the optimal configuration}


Since the UAVs move in a one-dimensional environment, the only decision to take at each step is either to go forward or backward along the only possible path. A reactive controller (Section \ref{sec:navigation}) keeps each UAV at the center of the tunnel.

We assume that all the UAVs are in a sub-optimal configuration and that they need to converge to the optimal one. To equalize signal quality, each UAV compares the quality of its connection with the previous and with the next UAV in the chain, and moves toward the UAV with which it has the worst link, as described in Algorithm (\ref{algo:simple}). By doing so, it improves the worst link quality and degrades the best one, moving them closer to equality. The speed at which the UAV moves is determined by the signal quality of its two neighboring links so that it does not move too fast (see \ref{proof:convergence}), and, consequently avoid oscillations. 


\begin{equation}
    \label{algo:simple}
    \begin{array}{l}
    \textrm{Let } s_{d} = s(x_{i}, x_{i-1}) - s(x_{i+1}, x_{i}), \\
    a_i \textrm{ moves to have}:
    \begin{cases}
        \textrm{if } s_d > 0, \\
        s(x_{i+1}, x_{i}) \textrm{ up by } \frac{s_d}{3}\\
        \textrm{if } s_d < 0, \\
        s(x_{i}, x_{i-1}) \textrm{ up by } \frac{s_d}{3}\\
    \end{cases}
    \end{array}
\end{equation}

\begin{claim}
    \label{result:convergence}
    The algorithm presented in (\ref{algo:simple}) converges to a configuration where all links' qualities are equal.
\end{claim}

\begin{claimproof}
    \label{proof:convergence}
    In \ref{proof:equality_constraints} we proved that unequal configurations were improved by moving by a small quantity $\epsilon$ an agent at the junction of two unequal links.
    Here, the algorithm does that in a distributed way for each agent seeing unequal links. We want to ensure that the simultaneous contributions from all agents still works the same way. We denote $\epsilon_i$ the movement of agent $a_i$, $s_i = s(x_{i+1}, x_i)$ the link quality before the movement and $\Tilde{s}_i = s(x_{i+1}+\epsilon_{i+1}, x_i+\epsilon_{i})$ the link quality after the movement.
    We assume that for each agent, $\epsilon_i$ is small enough that the contribution of $a_i$ to $\Tilde{s}_i$ is $\frac{s_{i-1}-s_{i}}{3}$.
    We look at one link $s_i$ that is one of the weakest assuming that $s_{i-1}$ is not one of the weakest links too. Agents from both sides will move to improve it, therefore $s_{i-1}$ and $s_{i+1}$ will be deteriorated. 
    We have that:
    \begin{equation}
    \begin{split}
        \Tilde{s}_{i-1} & = s_{i-1} + \frac{s_{i}-s_{i-1}}{3} + \frac{s_{i-2}-s_{i-1}}{3} \\
                        & = \frac{s_{i-1} + s_i + s_{i-2}}{3} \\
                        & > s_i \\
    \end{split}
    \end{equation}
    We can deduce the last inequality from the fact that $s_i < s_{i-1}$ and $s_i \leq s_{i-2}$ because $s_i$ is a weakest link and $s_{i-1}$ is not.
    We have similar result for $s_{i+1}$, ensuring that the deteriorated links are still better than the previously weakest link. Since the weakest links were improved, the new configuration is effectively better. For weakest links that have also weakest links as neighbors, nothing changes. In a non optimal configuration, at least one weakest link has a neighbor with a higher signal quality.
    
    We have an increasing sequence of configurations bounded by the optimal configuration, therefore the algorithm converges. As the only configuration for which the algorithm is stable is the equal links configuration - that is also optimal- we can deduce that this algorithm converges to the optimal configuration.
\end{claimproof}

\subsection{Computing $\epsilon_i$}
In order to ensure convergence, the algorithm needs to compute for each agent $a_i$ the distance to move $\epsilon_i$. This distance can be computed by solving the equation $s(x_{i+1}, x_i+\epsilon_i) - s(x_{i+1}, x_i) = \frac{s_d}{3}$.
Moreover, the proof of convergence of our algorithm states that if the agent $a_i$ moves exactly by $\epsilon_i$ at each step there is convergence. In fact, if it moves in the correct direction and that it moves at most $\epsilon_i$ then there is still convergence. Therefore, the exact computation of $\epsilon_i$ is not needed and an approximation can be sufficient.
\subsection{RSSI filtering with an optic flow-based Kalman filter}
The algorithm defined in (\ref{algo:simple}) makes all its decisions on measurements of the link quality, which is measured using the Received Signal Strength Indicator (RSSI). Unfortunately, these measures are noisy and not as reliable as we could hope. In particular, the signal quality is affected by many factors besides the emitter-receiver distance, including the orientation of the UAV, the current motor speed, the radio activity of the other UAVs, beams on the ceiling, etc. \cite{khuwaja_survey_2018,mcguire2019minimal}.

We therefore need to filter the RSSI to use it for our positioning algorithm. Using a simple moving average filter would not be sufficient as the estimation would have too much lag when the agents move towards or apart from each other. 

Our main insights are that (1) we know that the signal quality is likely to decrease when two UAVs move away from each other, and (2) the relative speed of two UAVs can be measured at no computational cost with miniature optic flow sensors \cite{ruffier2003bio} (commercial optical flow sensors derived from optical mouse sensors, like the Pixart PMW3901MB, weight a few mg and can work in low-light conditions \cite{mafrica2016optic}). 

We incorporate this simple model in a 1-dimensional Kalman filter to improve the signal quality estimation. To our knowledge, Kalman filters have never been used to improve the assessment of the signal quality in UAVs, in particular with optic flow sensors. Nevertheless, a few articles used RSSI and Kalman filters for localizing humans or phones by either assuming that the human does not move between two measurements \cite{7471364} or by combining RSSI and external sensors \cite{paul2009rssi}.

The filter considers that the signal quality decreases linearly with the relative velocity $u_k$:

\begin{eqnarray}
        r_k   &   = & r_{k-1} + A u_k + \epsilon   \\
        z_k   &   = & r_k + \delta                   
\end{eqnarray}

where $r_k$ is the the quantity to be filtered (the RSSI) and $u_k$ the system state (the relative speed between the agents) at iteration $k$. $\epsilon \sim \mathcal{N}(0,Q)$ is the intrinsic \emph{signal noise} and $A$ the impact of system state on the filtered quantity. $z_k$ is the observation of the estimated signal written as the true signal $r_k$ with a \emph{measurement noise} $\delta \sim \mathcal{N}(0,R)$.

The update of the Kalman filter occurs in two steps. Firstly, the \emph{prediction a priori} stage updates:

\begin{equation}
    \begin{split}
        \hat{r}_{k|k-1}   &   = \hat{r}_{k-1|k-1} + A u_k   \\
        P_{k|k-1}         &   = P_{k-1|k-1} + Q         \\
    \end{split}
\end{equation}

$\hat{r}_{k|k-1}$ is the prediction \emph{a priori} of the value of $r_k$ based only on the previous measures of it up to time $k-1$ and the current relative speed $u_k$. It has yet to take the new measurement $z_k$ into account. $\hat{r}_{k-1|k-1}$ is the estimation \emph{a posteriori} of the signal at time $k-1$. $P_{k|k-1}$ is variance \emph{a priori} of this estimation. The value of the constant $Q$ is an estimation of the \emph{signal noise} variance assumed to be known.

We finally compute the \emph{a posteriori} estimates with:

\begin{equation}
    \begin{split}
        K_k         &   = P_{k|k-1} \left( P_{k|k-1} + R \right)^{-1}\\
        \hat{r}_{k|k}   &   = \hat{r}_{k|k-1} + K_k \left( z_k - \hat{r}_{k|k-1} \right)\\
        P_{k|k}         &   = \left( I - K_k \right) P_{k|k-1}\\
    \end{split}
\end{equation}

with $K_k$ being the Kalman gain, and $\hat{r}_{k|k}$ and $P_{k|k}$ the estimation of signal and its variance \emph{a posteriori}.

In both simulated and real tests, we hand-tuned $A$ to minimize estimation noise as well as the \emph{lag} between the true RSSI and the estimation. 

The pseudo-code for the U-chain algorithm, which combines the movement policy and the Kalman filtering is displayed on Algorithm \ref{algo:opti}.




\begin{algorithm}[t]
    \caption{Main movement policy for each UAV $i \in R$ (head and base UAVs are excluded).} \SetAlgoLined 
    \SetAlgoLined 
    \label{algo:opti}
     \While{True} {
                $m_b = \textrm{measureRSSI}(i+1,i)$;\\
                $m_f = \textrm{measureRSSI}(i,i-1)$;\\
                $r_b, Q_b = \textrm{Kalman}_{A,R}(r_b, m_b,\dot{x}_{i+1} + \dot{x_{i}}, Q_b)$;\\
                $r_f, Q_f = \textrm{Kalman}_{A,R}(r_f, m_f,\dot{x_{i}} + \dot{x}_{i-1}, Q_f)$;\\
                $r_{\textrm{diff}} = r_b - r_f$;\\
                \uIf{$r_{\textrm{diff}} > \mathrm{T}$}{
                    $\Dot{x} = v(r_b, r_f)$\;
                }
                \uElseIf{$r_{\textrm{diff}} < - \mathrm{T}$}{
                    $\Dot{x} = v(r_b, r_f)$\;
                }
                \uElse {
                    $\Dot{x} = 0$\;
                } 
                setForwardVelocity($\Dot{x}$)\;
        }
\begin{itemize}
    \item The function $v$ can be deduced from the expression of $r$ (if known) so that by moving at velocity $v(r_{i+1, i},r_{i,i-1})$, $a_i$ changes $r_{i+1,i}$ by at most $\frac{r_{i,i-1} - r_{i+1,i}}{3}$ as stated in \ref{proof:convergence}.
    \item Threshold T (tolerance) is here to prevent numerical instability in the RSSI difference computations.
\end{itemize}

\end{algorithm}

\subsection{Exploration with a chain of UAVs} \label{sec:algo_exploration}
A straightforward use of algorithm \ref{algo:opti} is to move the first UAV (the head) forward in a tunnel and let the chain self-organize to maintain a stable connection. In that case, a new relay is launched when $s_{eq} > s_{\min}$, i.e. when the signal quality is too low. This can easily be implemented by having a set of idle UAVs waiting at the operator's position and making them take off when the other flying UAVs converged to an equal links configuration.
Assuming that each new take-off happens only when link quality stabilization has converged, $s_{eq}>s_{\min}$, it means the head will advance the farthest possible before using a new UAV. As a consequence, the minimal number of relays will be used for a given head position $x_t$.

If no more relay UAVs are available to take off, the chain still equalizes the links' qualities but may go under $s_{\min}$. Additionally, the head UAV may move too fast and lose connection while exploring a new area (e.g. because of a sudden bad propagation due to the local environment). A connection may still be possible but is not guaranteed. Therefore, all UAVs move automatically backwards while the connection to its predecessor is too weak or lost.

\begin{figure}
    \centering
    \includegraphics[width=.90\linewidth]{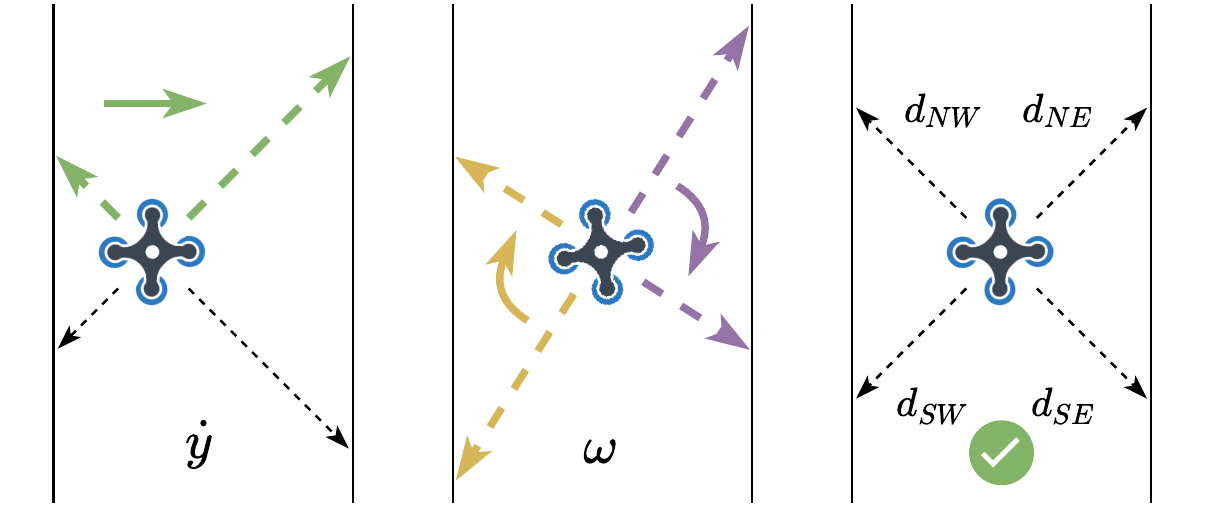}
    \caption{\textbf{Main centering policy} (equation \ref{eq:navigation}). The quadrotors use the two front distance sensors ($d_{NW}$ and $d_{NE}$) to stay at the center of the corridor, and the difference between the sensors of the same side (e.g., $d_{NW}$ and $d_{SW}$) to control the yaw.}
    \label{fig:navigation}
    \vspace{-2em}
\end{figure}

\subsection{Centering policy} \label{sec:navigation}
    Our UAVs need to autonomously follow the explored tunnel, which can be achieved with a few miniature time-of-flight sensors (we use VL53L1x by ST Electronics). To do so, a reactive controller runs independently from the positioning algorithm to center each drone and follow the corridor turns. This policy uses four diagonal distance sensors to compute both the yaw and the lateral velocities (perpendicular to the wall). More precisely, at each iteration, in parallel with the calculation of $\dot{x}$ (Algorithm~\ref{algo:opti}):

    \begin{equation}
        \label{eq:navigation}
        \begin{cases}
            \dot{y} = C_t \cdot (d_{NW} - d_{NE}) \\ 
            \omega \, = C_r \cdot (d_{NW} - d_{SW}) + C_r \cdot (d_{SE} - d_{NE})
        \end{cases}
    \end{equation}
    
    where $\dot{y}$ is the lateral velocity (the correction), $d_{NW}, d_{NE},d_{NW}, d_{SW}$ are the four distance sensors (see Fig. \ref{fig:navigation}), and both $C_t$ and $C_r$ are used-defined constants. Essentially, this policy uses the two front sensors to center in the corridor (the UAV is at the center of the corridor when both sensors have equal values), and the difference between the sensors from the same side for the yaw (the UAV is aligned with the walls when both sensors return the same value).

    While surprisingly effective, this reactive policy is not always sufficient, in particular for 90 degrees turns. In this situation, the distance of one of the front sensors suddenly increases (e.g. $d_{NE}$ for a right turn), which creates an large distance difference between the two front sensors, and results in a dangerous $y$ correction.
   
    To mitigate this issue, we calculate the ratio of the distances of each pair of sensors that sense the left and right sides of the UAVs. If the difference percentage goes beyond a fixed value (40\% after calibration tests), we consider that the wall is invalid ignore it in the centering policy. When this happens, the drone automatically follows the remaining wall at a fixed distance by applying (example when the right wall is lost):
   
   \begin{equation}
        \label{eq:navigation_1wall}
        \begin{cases}
            \dot{y} = C_t \cdot (d_{NW} - D) \\ 
            \omega \, = 2 \cdot C_r \cdot (d_{NW} - d_{SW})
        \end{cases}
    \end{equation}

    with $D$ being a preset distance. This additional policy makes it possible to explore wide rooms by following one side of the room, in addition to center in a tunnel.

\begin{figure*}
    \centering
    \includegraphics[width=\linewidth]{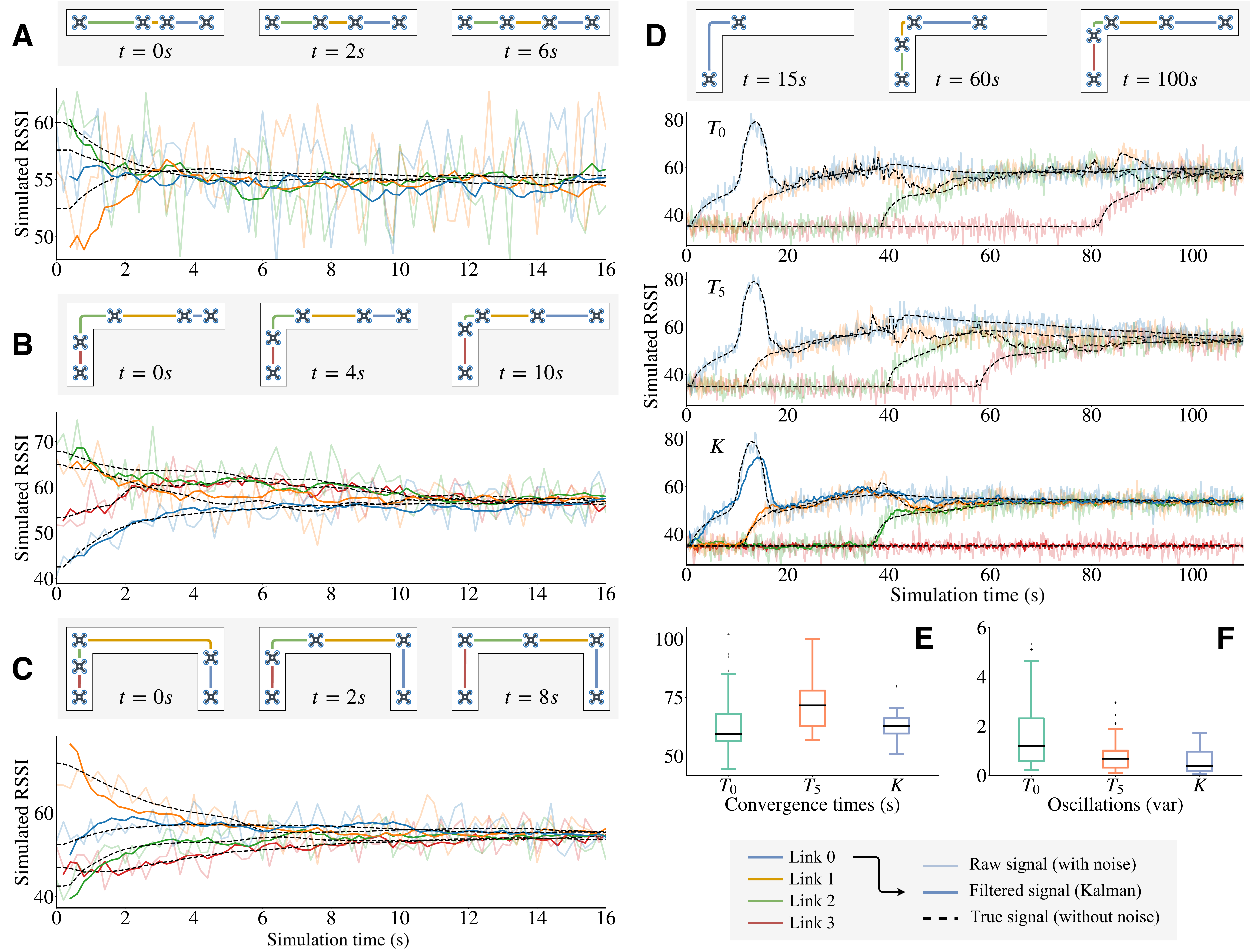}
    \caption{\textbf{Simulation results.} For each figure, the true signal (dashed line) is computed using equation \ref{eq:rssi} but without the Gaussian noise (this is unknown to our algorithm); the raw signal (light solid line) is the input of the U-Chain algorithm; and the filtered signal is the result of the Kalman filter used to take the movement decisions. \textbf{A-C.} Convergence of Algorithm \ref{algo:opti} in different environments (shown in the corresponding map thumbnails). \emph{The UAVs place themselves at ``strategic'' positions}. \textbf{D.} Comparison between three different signal processing methods: $T_0$ for no processing, $T_5$ for a tolerance of local signal difference of 5 and $K$ for using a Kalman-filtered estimation of RSSI. \textbf{E.} Comparison of the convergence times for 30 independent replicated runs on each of these three methods. \textbf{F.} Variance of the UAV positions after convergence (30 replicates).
    \emph{We observe that $K$ gives a short convergence time (compared to $T_5$) and less oscillation amplitude (compared to $T_0$).}}
    \label{fig:sim_results}
    \vspace{-1.0em}
\end{figure*}

\section{Experimental results} \label{sec:results}

    \subsection{Simulation}
        \subsubsection{Main assumptions}
        \begin{itemize}
            \item The internal decision loop of each agent runs at 5Hz to ensure that the algorithm will be easy to implement in the Crazyflies' micro-controller. 
            \item The UAVs communicate with simulated packet transmissions that are possible only if $s(x_1, x_2) > s_{min}$.
            \item The environment is 2-dimensional.
            \item The distance sensors are computed according to the environment and the UAVs follow the navigation policy (section \ref{sec:navigation}).
        \end{itemize}

        \subsubsection{Signal propagation model}
            We model the RSSI according to models given by \cite{vinogradov_tutorial_2019}. The chosen signal loss estimation is expressed as:
            
            \begin{equation}
            \label{eq:rssi}
                RSSI(x_1, x_2) = 10 \times \alpha(x_1, x_2) \times \log (|x_1 - x_2|) + \epsilon
            \end{equation}
            
            with $\alpha$ being the environment's attenuation factor. When fixed, this expression corresponds to the \emph{path~loss} caused by air attenuation and is the main cause of signal degradation. Then, we include the \emph{shadowing} phenomenon due to obstacles blocking the line of sight between the communicating agents by varying $\alpha$, based on the environment. The value varies between 2 and 6 according to the amount of walls the segment between the agents passes through. $\epsilon \sim \mathcal{N}(0,B)$ is a Gaussian noise with a variance of 3.
            
            Please note that RSSI$(x_1,x_2)$ is defined as an increasing function, whereas we assumed so far that the signal quality decreases with the distance. As a consequence we choose $s(x_1, x_2) = -RSSI(x_1, x_2)$.
            Finally, we introduce a fixed 20\% chance of losing a packet which is consistent with our experience with the Crazyflies.
            

        \begin{figure*}
        \centering
        \label{img:real_circuit}
        \label{graph:real_graph}
        \includegraphics[width=.85\columnwidth, valign=t]{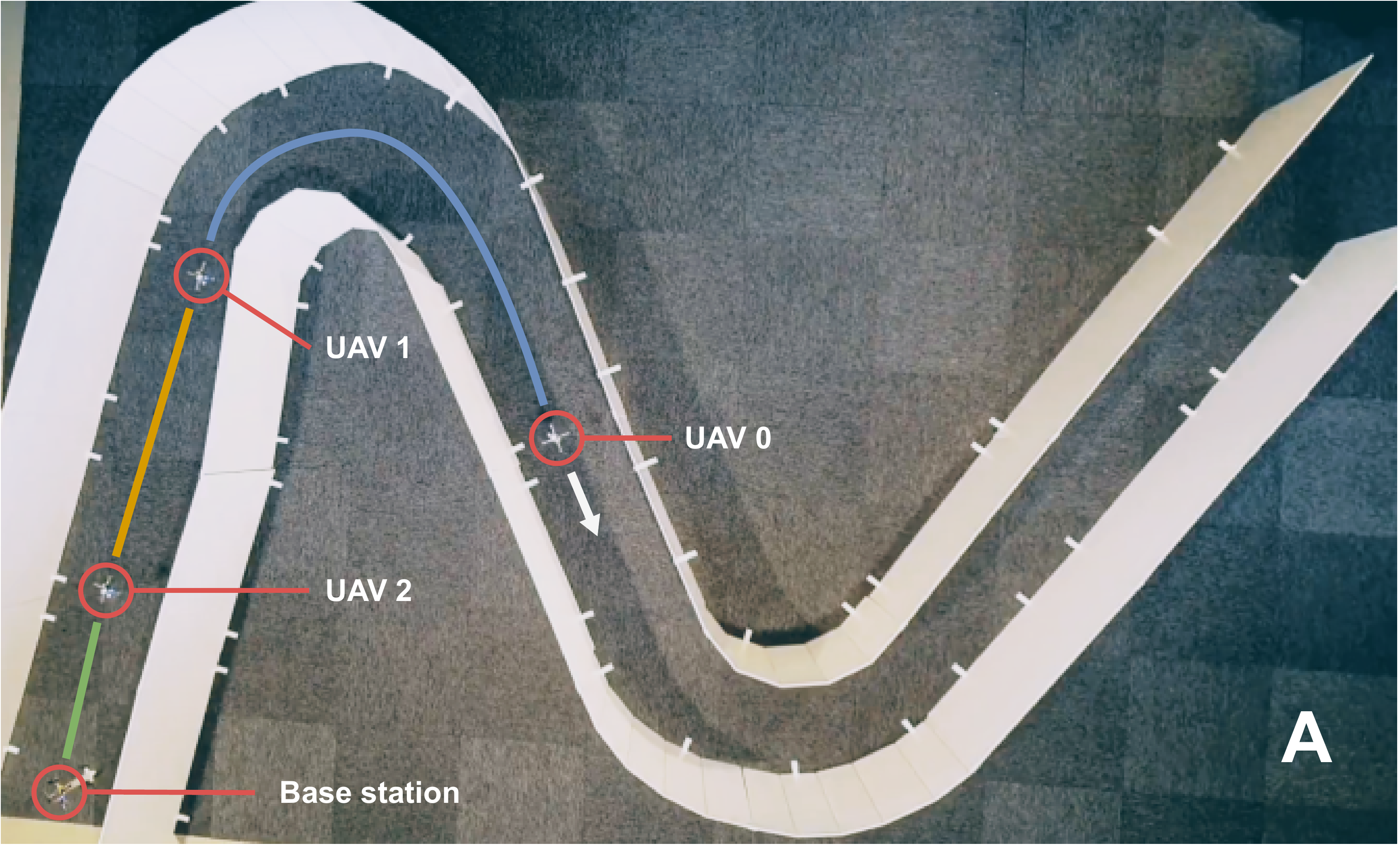}
        \hfill
        \includegraphics[width=0.90\columnwidth, valign=t]{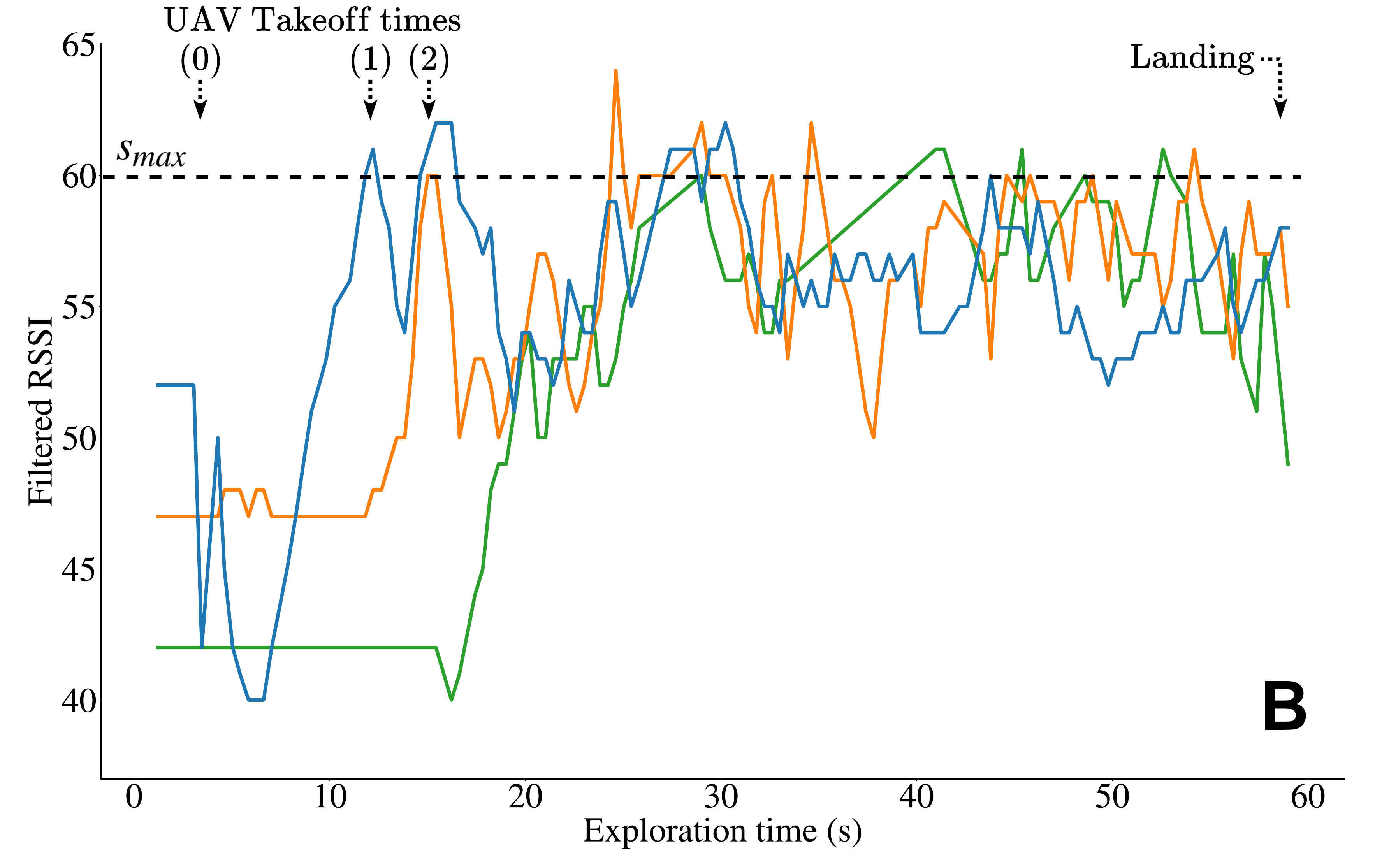}
    \caption{\textbf{Experiments with 3 Crazyflie 2.1 quadrotors and a base station.} The robots use an optic flow sensor and 5 time-of-flight sensor (altitude and centering). They communicate with a custom peer to peer protocol. \textbf{A.} Environment used for the experiments. \textbf{B.} Qualities of the radio links (RSSI) during exploration. \emph{A video is available with the submission.}}
    \label{fig:real_result}
        \vspace{-2em}
\end{figure*}

        \subsubsection{Results}
                We firstly checked that the chain successfully converges to stable positions and equalizes the signal quality when $n$ UAVs are already flying in the air. To do so, we positioned the UAVs at random positions in the corridor for three different environments while fixing the position of the head. The results (Fig. \ref{fig:sim_results}-A-C) show that the UAVs always converge to the desired solution, in spite of the noise on RSSI measurements. We confirmed this result by randomly generating 30 initial positions for 30 independent runs for each of the three environments: 100\% of the runs converged to positions that equalize the RSSI.


                We then evaluated the usefulness of the Kalman filter when the head drone is moving. We simulated a real exploration by placing all UAVs at the entrance of the corridor and launching the first UAV at the start of the simulation. The head then moves forward at a constant speed ($0.2$ meters per second) before stopping after 50 seconds of exploration. 
                
                When using the raw signal as the entry for algorithm \ref{algo:opti} (Fig.~\ref{fig:sim_results}-D, algorithm $T_0$), the system successfully converges as expected (Fig.~\ref{fig:sim_results}-E). However, a single abnormally high value of RSSI caused by the noise can trigger the launch of an additional drone (see the red curve at 80 seconds) that is not needed to keep the true signal quality above $s_{\min}$. Moreover, the variance of the UAV positions after reaching the final positions is particularly high (Fig.~\ref{fig:sim_results}-F) which may cause stability problems when testing the system on real micro UAVs. Figures E and F only take the first three links' qualities into account in order to ignore the non-repeatable launch of the forth UAV.
                
                A simple way of trying to reduce these oscillations is to introduce a tolerance around the local quality equilibrium of each UAV before applying any movement (algorithm $T_5$). This indeed reduces the oscillations after convergence but unfortunately increases the converge time, as small position corrections near convergence only happen when the noise makes the RSSI measurement exceed the tolerance value.
                
                The Kalman filter improves the system performance on those two factors. The estimation greatly reduces the noise of the signal used in algorithm \ref{algo:opti} which leads to both a short convergence time and less oscillation.



            
            

    \subsection{Real-world experiment}


            We designed a circuit that resembles a real corridor or pipe and respects the monotonic propagation constraint. The walls are made of thin paper boards for installation simplicity, and therefore do not block the signals in any way. However, they prevent the UAVs from taking the shortest path, which means that the signal quality follows a complex non-linear function along the tunnel.  
            
            The micro UAVs used in our experiments are off-the-shelf commercial micro drones (Crazyflie 2.1 by Bitcraze) with two sensor add-ons (called ''decks''). The first deck provides four time-of-flight distance sensors that detect obstacles up to two meters (the detection angle is 27 degrees), which are used by the centering policy (section \ref{sec:navigation}). The second deck provides a fifth time-of-flight distance sensor that points downward, which is used to stabilize the UAV vertically. The same deck embeds an optical flow sensor, which helps stabilizing the UAV (by avoiding to drift) and provides the data for the ground speed estimate used in our algorithm (Sec.\ref{sec:algorithm}).
            
            We added drone-to-drone (peer to peer) communication, as the original firmware only supported packet transmissions between the micro UAVs and a computer. The source code of the upgraded firmware code is available online (\url{https://github.com/resibots/crazyflie-firmware/}, branch 'cavemod').
            
            In the experiments, all UAVs are initially positioned at the start of the corridor (Fig. \ref{fig:real_result}-A) and are ready to take off when needed, except the last drone, which acts as as a fixed base station. We launch the first UAV and manually control it with the keyboard keys (the pilot can only go forward or backward --- the actual stabilization and centering is autonomous). The other UAVs then apply all presented algorithms in this paper to launch and relay the signal when needed.
            
            Overall, the system performs similarly to the simulations (Fig. \ref{fig:real_result}-B): all relays take off when the last active link reaches the fixed limit, and the UAVs successfully equalize the 3 links' qualities until the end of the exploration. A video is available as supplementary material.

\section{CONCLUSION}
By combining a reactive movement policy with a Kalman filter, the U-Chain algorithm can coordinate a chain of UAVs in a tunnel to maintain a high-quality connection while being light enough to be embedded on miniature UAVs. Since only the signal's qualities are considered, the chain of UAVs adapts to any unexpected signal propagation, turns, radio perturbations, etc. While we did not consider intersections in this work, it is straightforward to store and transmit the direction choice made by the pilot who is controlling the head UAV, provided that a crude approximate of the position $x$ in the tunnel is available (so that the following UAV can take the same decision).

The main hypothesis of the present work is that the quality of the signal decreases monotonically with the distance between an emitter and a receiver. This hypothesis is reasonable in an underground environment in which the signal never pass through walls: the signal is very unlikely to get better when the UAVs progress into a tunnel. However, this is often not the case in more general indoor environments, and in particular in modern buildings in which the radio signal can often pass through walls: in these cases, the signal can improve if there exists a path through the walls that is shorter than the physical path available (e.g., a door or a corridor). Future work will attempt to relax the monotony hypothesis while ensuring the convergence properties, in particular when the head UAV is moving.


\bibliographystyle{IEEEtran}
\bibliography{IEEEabrv,references}
\end{document}